\documentclass[11pt,a4paper]{article}
\usepackage{times}
\usepackage{latexsym}
\usepackage{url}
\usepackage{hyperref}
\usepackage[all]{hypcap}
\usepackage{graphicx}
\usepackage{amsmath}
\usepackage{amssymb}
\usepackage{xspace}
\usepackage{fixfoot}
\usepackage{multirow}
\usepackage{ marvosym }
\usepackage{calc}
\usepackage{booktabs}
\usepackage{ textcomp }
\usepackage{mathtools}
\usepackage{stfloats}
\usepackage{naacl2021}

\title{Exploring the Relationship Between Algorithm Performance, Vocabulary, and Run-Time in Text Classification}

\author{Wilson Fearn \\
  Brigham Young University \\
  \texttt{wilson.fearn@gmail.com} \\\And
   Orion Weller \\
 Brigham Young University \\
  \texttt{oweller@byu.edu} \\\And
  Kevin Seppi \\
  Brigham Young University \\
  \texttt{kseppi@byu.edu} \\}
  
\begin{document}

\maketitle

\begin{abstract}
Text classification is a significant branch of natural language processing, and has many applications including document classification and sentiment analysis. 
Unsurprisingly, those who do text classification are concerned with the run-time of their algorithms, many of which depend on the size of the corpus' vocabulary due to their bag-of-words representation.
Although many studies have examined the effect of preprocessing techniques on vocabulary size and accuracy, none have examined how these methods affect a model's run-time.
To fill this gap, we provide a comprehensive study that examines how preprocessing techniques affect the vocabulary size, model performance, and model run-time, evaluating ten techniques over four models and two datasets.
We show that some individual methods can reduce run-time with no loss of accuracy, while some combinations of methods can trade 2-5\% of the accuracy for up to a 65\% reduction of run-time.  Furthermore, some combinations of preprocessing techniques can even provide a 15\% reduction in run-time while simultaneously improving model accuracy.\footnote{Our code and results are publicly available at \url{https://github.com/wfearn/preprocessing-paper}}
\end{abstract}

\section{Introduction}
\label{sec:intro}
With the increasing amount of text data available, text analysis has become a significant part of machine learning (ML). Many problems in text analysis use ML methods to perform their task, ranging from classical problems like text classification and topic modeling, to more complex tasks like question answering. Although neural networks have become increasingly common in the research field, many industry NLP problems can be well served by less complex but more efficient and explainable models, such as Support Vector Machines (SVMs) or K-Nearest Neighbors (K-NN).

We focus on the text classification problem, where the dominant approach to using these non-neural models is to first calculate the number of unique terms in the dataset (the \textit{vocabulary}, size $V$) and encode each instance of the dataset into a bag-of-words (BoW) representation \cite{joachims1998text,Zhang2010UnderstandingBM}. This results in a high-dimensional vector of size $V$ that indicates whether each given word of the vocabulary was used in this instance.

However, the vanilla approach to the BoW representation can lead to sub-par performance, as shown by numerous studies that have examined how preprocessing techniques affect the BoW w.r.t. performance and vocabulary size. These studies have examined this representation in fields such as information retrieval
\cite{chaudhari2015preprocessing,patil2013novel,beil2002frequent,senuma_2011}, text classification \cite{Yang1997ACS,caragea2012combining,uysal2014impact,vijayarani2015preprocessing,kumar2018classification,hacohen2020influence,symeonidis_effrosynidis_arampatzis_2018} and topic modeling \cite{schofield2016comparing,blei2003latent}.  They suggest a myriad of preprocessing techniques that could improve performance, ranging from choosing features that have high mutual information, low frequency, or simply remove punctuation.

Another related problem of the BoW representation is that this sparse high-dimensional vector does not scale well to datasets with large vocabularies. As preprocessing techniques help contribute to a reduced vocabulary, they should also help alleviate this scaling problem, at least according to folklore. However, to the best of our knowledge, no previous study of preprocessing techniques have examined how they contribute to reduced run-time costs, leading to uncertainty about what these techniques do to mitigate the computational complexity in practice.

\begin{figure}[t!]
    \centering
    \includegraphics[trim=25 20 0 0,width=0.48\textwidth]{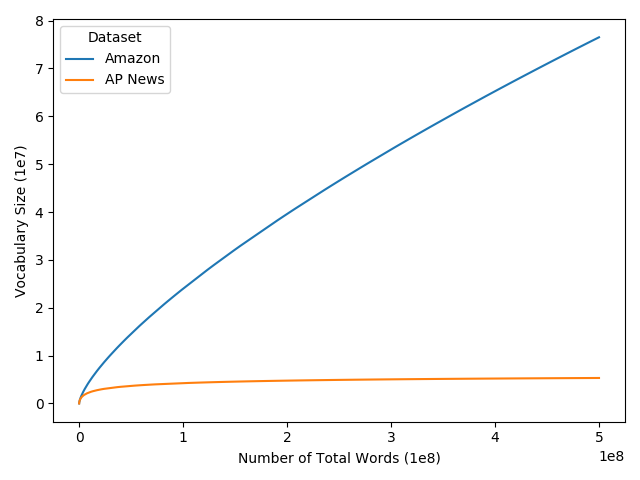}
    \caption{Comparing vocabulary size (in millions) vs the total number of words (in 10s of millions) for the AP News and Amazon corpora. Note that the vocabulary size of AP News w.r.t. the number of documents plateaus much faster than the noisier Amazon corpus.}
    \label{fig:vocab_curve}
\end{figure}

To remedy this, we analyze how these preprocessing methods affect not only vocabulary size and performance, but also how they affect training and inference time. To do this, we contribute a comprehensive analysis of 10 different preprocessing methods applied to four machine learning models, evaluated on two datasets with widely varying vocabularies (Figure~\ref{fig:vocab_curve}). 

Our results show that the individual preprocessing methods provide widely different effects on run-time, with some methods (i.e. rare word filtering and stopword removal) providing significant run-time reductions without losing any performance. We also show that some combinations of preprocessing methods both improve performance and reduce run-time.

\section{Experimental Setup}
\label{sec:setup}
\paragraph{Datasets} To see how preprocessing affects run-time, we examine two datasets (in English): the Amazon \cite{he2016ups}\footnote{\url{http://jmcauley.ucsd.edu/data/amazon/}} and AP News corpora \cite{macintyre1998north}. These datasets were chosen because of the wide disparity between their vocabularies. The Amazon corpus comes from user product reviews and contains a much higher vocabulary relative to the number of documents, due to its noisy text. The AP News corpus contains professionally-edited news articles and its vocabulary plateaus much faster than the Amazon corpus (Figure~\ref{fig:vocab_curve}). We perform sentiment analysis on Amazon and year classification on AP News and report scores with the accuracy metric. We note that we also computed the F1 score alongside accuracy and found our results to be similar; thus we report accuracy since it is easier to understand.

\begin{table*}[t!]
\small
\centering
\begin{tabular}{l|lrrrr}
\toprule
                   Group &                     Method &                  Vocab Size~$\downarrow$ &                   Train Time~$\downarrow$ &                    Test Time~$\downarrow$ &                      Accuracy~$\uparrow$ \\
\midrule
                        &                       stop &          99.8  \textpm\ 0.2 &  \textbf{69.5}  \textpm\ 1.4 &           79.0  \textpm\ 3.5 &            97.4  \textpm\ 2.2 \\
                        &                       rare &  \textbf{1.0}  \textpm\ 0.0 &           80.6  \textpm\ 3.0 &  \textbf{70.3}  \textpm\ 3.2 &   \textbf{99.3}  \textpm\ 2.8 \\
                        &                        seg &          24.6  \textpm\ 0.2 &           93.7  \textpm\ 2.4 &           80.3  \textpm\ 2.5 &  \textbf{100.6}  \textpm\ 1.6 \\
                        &                      spell &          57.8  \textpm\ 0.2 &           95.1  \textpm\ 2.6 &           89.7  \textpm\ 2.6 &   \textbf{99.4}  \textpm\ 2.3 \\
     Individual Methods &                       hash &          10.1  \textpm\ 0.0 &           97.1  \textpm\ 4.0 &           75.7  \textpm\ 4.0 &   \textbf{99.2}  \textpm\ 1.1 \\
                        &                    nopunct &          61.9  \textpm\ 0.2 &           97.5  \textpm\ 2.2 &           89.5  \textpm\ 2.0 &  \textbf{100.7}  \textpm\ 1.6 \\
                        &                       stem &          81.7  \textpm\ 0.4 &           97.8  \textpm\ 2.0 &           95.0  \textpm\ 2.6 &   \textbf{99.8}  \textpm\ 1.0 \\
                        &                      lower &          88.7  \textpm\ 0.3 &          101.7  \textpm\ 7.5 &          100.1  \textpm\ 6.6 &   \textbf{99.1}  \textpm\ 3.0 \\
                        &                       nrem &          96.2  \textpm\ 0.7 &          101.7  \textpm\ 4.0 &          100.7  \textpm\ 5.3 &   \textbf{99.7}  \textpm\ 1.2 \\
                        &                      lemma &          98.1  \textpm\ 0.5 &          102.2  \textpm\ 5.3 &          101.5  \textpm\ 5.1 &  \textbf{100.3}  \textpm\ 1.1 \\
\midrule

                        &   spell+seg+nrem+stop+rare &  \textbf{0.8}  \textpm\ 0.0 &  \textbf{44.6}  \textpm\ 1.0 &           56.8  \textpm\ 1.1 &            95.4  \textpm\ 2.0 \\
                        &                  stop+rare &           0.9  \textpm\ 0.0 &  \textbf{46.5}  \textpm\ 3.5 &           44.5  \textpm\ 2.0 &   \textbf{99.8}  \textpm\ 0.8 \\
 Lowest Train/Test Time &   spell+seg+nrem+stop+hash &           7.6  \textpm\ 0.0 &           53.9  \textpm\ 2.0 &  \textbf{39.6}  \textpm\ 1.4 &   \textbf{97.7}  \textpm\ 2.6 \\
                        &        spell+seg+nrem+stop &          14.1  \textpm\ 0.0 &           54.2  \textpm\ 1.6 &           50.9  \textpm\ 2.3 &   \textbf{97.6}  \textpm\ 2.2 \\
                        &  spell+seg+nrem+stop+lemma &          11.9  \textpm\ 0.0 &           55.1  \textpm\ 0.9 &           50.1  \textpm\ 1.7 &            97.6  \textpm\ 1.3 \\
\midrule
                        &               nopunct+rare &  \textbf{0.9}  \textpm\ 0.0 &  \textbf{82.6}  \textpm\ 1.9 &  \textbf{70.2}  \textpm\ 1.7 &  \textbf{101.0}  \textpm\ 1.8 \\
                        &    lower+nopunct+nrem+rare &  \textbf{0.9}  \textpm\ 0.0 &  \textbf{87.1}  \textpm\ 5.5 &           88.8  \textpm\ 5.1 &           101.1  \textpm\ 0.3 \\
       Highest Accuracy &         lower+nopunct+rare &  \textbf{0.9}  \textpm\ 0.0 &           86.1  \textpm\ 2.7 &           86.7  \textpm\ 2.2 &  \textbf{101.3}  \textpm\ 0.6 \\
                        &                   seg+rare &  \textbf{0.9}  \textpm\ 0.0 &  \textbf{86.3}  \textpm\ 5.8 &  \textbf{73.3}  \textpm\ 3.9 &  \textbf{101.4}  \textpm\ 1.5 \\
                        &             spell+seg+rare &  \textbf{0.9}  \textpm\ 0.0 &           89.6  \textpm\ 5.6 &           88.4  \textpm\ 5.3 &  \textbf{101.8}  \textpm\ 0.5 \\
\bottomrule
\end{tabular}

\caption{Effect of preprocessing techniques on Amazon. Scores are the relative performance of each method over the \textit{no preprocessing} baseline (e.g. stopword removal takes only 69.5\% of the baseline's training time). Results are the average (and std) relative performance of the four models, across the five dataset seeds. Bold indicates statistical similarity to the best score, from a two-sample t-test with $\alpha=0.05$. For brevity, this table only includes individual methods and those with the highest accuracy or lowest train/test time. All results are in Appendix~\ref{app:full}.\label{tab:amazon}}
\end{table*}

To test the effect of document size on preprocessing, we sampled various-sized\footnote{5k, 10k, 20k, 30k, 40k, 50k, and 100k instances} datasets from the original corpus and ran our analysis on each, sampling 5 different times with differing random seeds.\footnote{Although the Amazon corpus contains many more documents, we keep our sampling consistent with the AP News corpus, as AP News has only 600k instances.} However, we found that our results were nearly identical across the differing corpus sizes and thus, only report numbers for the 100k size.

\paragraph{Preprocessing Methods}
We analyze 10 different methods (with their shortened names in parenthesis): lowercasing (lower), rare word filtering (rare), hashing (hash), punctuation removal (nopunct), stopword removal (stop),
number removal (nrem), word stemming (stem), lemmatization (lemma), spelling correction (spell), and word segmentation (seg). We choose these methods because of their prevalence in previous work \cite{symeonidis_effrosynidis_arampatzis_2018,kumar2018classification,hacohen2020influence} and their use in industry \cite{li2013determinants,sanchez2014text}.

Due to the exponential number of possible preprocessing combinations, we run all individual methods but restrict the search space of combinations of these methods. For rare word filtering and word hashing, we first conduct experiments for 9 different levels of filtering individually, using only the best level in future combinations with other methods. Results for all levels of filtering and hashing are in Appendices \ref{app:rare} and \ref{app:hash}. We then conduct experiments for all 24 combinations of spelling correction, word segmentation, number removal, and stopword removal, using the best outcome (the pipeline of all four) to combine with other methods. We note that while this is not an exhaustive search of all combinations, our analysis includes the standard preprocessing pipelines as well as many more.

\paragraph{Models}
We use Scikit-Learn \cite{scikit-learn} for three of the base algorithms, including K-NN \cite{altman1992introduction}, Naive Bayes \cite{rish2001empirical}, and the Support Vector Machine (SVM, \cite{suykens1999least}). We also employ Vowpal Wabbit \cite{langford2007vowpal,2010arXiv1011}, due to its strong performance and frequent use in industry. All models use default hyperparameters and our document representations use the BoW representation, consisting of a sparse vector format. These four models provide a wide range of algorithms that might be used, allowing us to show how preprocessing methods generalize across models.

\paragraph{Compute}
All experiments were performed using 14-core Intel Broadwell
processors running at 2.4GHz with 128GB of DDR4 2400 MT/s RAM.

\section{Results}
\label{sec:results}
We format our results relative to the algorithm with no preprocessing, to easily show how preprocessing changes this baseline performance. We first run each algorithm with no preprocessing, measuring the run-time, vocabulary size, and accuracy. We then report the scores of each preprocessing pipeline relative to the algorithm's baseline (e.g. a model with preprocessing that scores 75\% of the no-preprocessing baseline's accuracy has a relative accuracy of 0.75).

As the cross product of the number of methods vs. the number of models is still far too large to include in this paper, we show the average of each model's relative proportion to its respective baseline performance.\footnote{We first compute each algorithm's relative score to its baseline (e.g. SVM with rare word filtering vs SVM with no preprocessing) and then take the average of the models for that method (e.g. average the relative performance of rare word filtering on models \{K-NN, Naive Bayes, SVM, and Vowpal Wabbit\} for the final score for rare word filtering).} This aggregation shows us the average relative performance across the four models, helping us generalize our results to be model-independent. For full tables detailing specific model results, see Appendix \ref{app:full}. Bold scores in tables indicate statistical similarity to the best score in the column (two-sample t-test, $\alpha=0.05$).

\begin{table*}[t!]
\small
\centering
\begin{tabular}{l|lrrrrrrr}
\toprule
                   Group &                     Method &                  Vocab Size~$\downarrow$ &                   Train Time~$\downarrow$ &                    Test Time~$\downarrow$ &                      Accuracy~$\uparrow$ \\
\midrule
                        &                           rare &   \textbf{0.1}  \textpm\ 0.0 &  \textbf{51.4}  \textpm\ 1.2 &  \textbf{59.1}  \textpm\ 2.2 &   \textbf{99.8}  \textpm\ 2.0 \\
                        &                           stop &           99.5  \textpm\ 0.3 &           82.5  \textpm\ 2.9 &           86.4  \textpm\ 2.0 &   \textbf{99.0}  \textpm\ 1.4 \\
                        &                           hash &           32.8  \textpm\ 0.0 &           98.5  \textpm\ 4.7 &           84.7  \textpm\ 4.0 &   \textbf{99.5}  \textpm\ 1.7 \\
                        &                          spell &           65.6  \textpm\ 0.2 &           98.6  \textpm\ 4.7 &           95.2  \textpm\ 7.4 &   \textbf{99.7}  \textpm\ 0.9 \\
     Individual Methods &                          lower &           92.3  \textpm\ 0.3 &           98.9  \textpm\ 1.9 &           97.6  \textpm\ 3.5 &   \textbf{99.8}  \textpm\ 1.6 \\
                        &                           stem &           82.7  \textpm\ 0.4 &           99.1  \textpm\ 5.2 &           95.2  \textpm\ 4.0 &  \textbf{100.1}  \textpm\ 1.4 \\
                        &                            seg &           47.5  \textpm\ 0.2 &           99.5  \textpm\ 2.4 &           88.5  \textpm\ 2.9 &  \textbf{100.3}  \textpm\ 1.3 \\
                        &                           nrem &           89.8  \textpm\ 0.4 &           99.8  \textpm\ 3.9 &           98.5  \textpm\ 5.0 &   \textbf{99.2}  \textpm\ 1.1 \\
                        &                        nopunct &           65.6  \textpm\ 0.2 &           99.9  \textpm\ 4.6 &           92.9  \textpm\ 4.9 &   \textbf{99.6}  \textpm\ 1.4 \\
                        &                          lemma &           97.4  \textpm\ 0.3 &          100.5  \textpm\ 1.2 &           98.6  \textpm\ 1.6 &   \textbf{99.6}  \textpm\ 1.7 \\
\midrule
                        &       spell+seg+nrem+stop+rare &   \textbf{0.1}  \textpm\ 0.0 &  \textbf{29.2}  \textpm\ 0.5 &  \textbf{49.0}  \textpm\ 0.8 &   \textbf{99.3}  \textpm\ 1.7 \\
                        &        spell+nopunct+nrem+stop &           39.9  \textpm\ 0.1 &           71.0  \textpm\ 1.8 &           69.4  \textpm\ 1.8 &  \textbf{100.1}  \textpm\ 1.0 \\
 Lowest Train/Test Time &  spell+nopunct+nrem+stop+lemma &           36.2  \textpm\ 0.1 &           71.3  \textpm\ 1.3 &           68.4  \textpm\ 1.4 &   \textbf{99.1}  \textpm\ 1.6 \\
                        &            spell+seg+nrem+stop &           29.3  \textpm\ 0.0 &           72.1  \textpm\ 1.8 &           65.4  \textpm\ 3.2 &  \textbf{100.1}  \textpm\ 1.4 \\
                        &   spell+nopunct+nrem+stop+stem &           29.6  \textpm\ 0.1 &           72.4  \textpm\ 1.2 &           68.2  \textpm\ 1.8 &   \textbf{98.7}  \textpm\ 1.8 \\
\midrule
                        &       spell+seg+nrem+stop+stem &  \textbf{19.3}  \textpm\ 0.1 &           74.2  \textpm\ 2.3 &  \textbf{66.9}  \textpm\ 1.5 &   \textbf{99.7}  \textpm\ 1.6 \\
                        &        spell+nopunct+nrem+stop &           39.9  \textpm\ 0.1 &  \textbf{71.0}  \textpm\ 1.8 &           69.4  \textpm\ 1.8 &  \textbf{100.1}  \textpm\ 1.0 \\
       Highest Accuracy &            spell+seg+nrem+stop &           29.3  \textpm\ 0.0 &  \textbf{72.1}  \textpm\ 1.8 &  \textbf{65.4}  \textpm\ 3.2 &  \textbf{100.1}  \textpm\ 1.4 \\
                        &       spell+seg+nrem+stop+hash &           19.8  \textpm\ 0.0 &  \textbf{73.1}  \textpm\ 2.7 &  \textbf{66.6}  \textpm\ 3.4 &  \textbf{100.2}  \textpm\ 1.5 \\
                        &   lower+nopunct+nrem+stop+stem &           39.4  \textpm\ 0.4 &           75.5  \textpm\ 2.4 &           72.7  \textpm\ 3.0 &  \textbf{100.3}  \textpm\ 1.0 \\
\bottomrule
\end{tabular}

\caption{Effect of preprocessing techniques on AP News. Scores are the relative performance of each method over the \textit{no preprocessing} baseline (e.g. stopword removal takes only 82.5\% of the baseline's training time). Results are the average (and std) relative performance of the four models, across the five dataset seeds. Bold indicates statistical similarity to the best score, from a two-sample t-test with $\alpha=0.05$. For brevity, this table only includes individual methods and those with the highest accuracy or lowest train/test time. All results are in Appendix~\ref{app:full}.\label{tab:ap_news}}
\end{table*}

\paragraph{Individual Techniques}
We see results for the Amazon corpus in Table~\ref{tab:amazon} and for the AP News corpus in Table~\ref{tab:ap_news}. On Amazon, each individual preprocessing method performs statistically similar to the baseline's accuracy, while three algorithms (stopword removal, rare word filtering, and word segmentation) also provide a moderate decrease (20-30\%) in train and test time. Rare word filtering and stopword removal are effective across both corpora (with rare word filtering being even more effective on AP News, reducing the training time in half), while the other methods do not significantly impact either train-time or accuracy on AP News. We hypothesize that these techniques are more effective on the AP corpus because of its much smaller (and less varied) vocabulary.

\paragraph{Combination Techniques}
The combination techniques also show a mild impact on accuracy, with most methods on both corpora performing statistically similar to the baseline. On the Amazon corpus, a handful of methods trade 2-5\% of accuracy for up to a 65\% reduction in training and testing time (``Lowest Train/Test Time" section in Table~\ref{tab:amazon}). Those that do not reduce accuracy (such as stop+rare) can still reduce the training and testing time by up to 55\%.  We see in the ``Highest Accuracy" section that some methods (i.e. spell+seg+rare, etc.) can even improve performance by almost 2\% while also reducing run-time by 10-15\%. Similarly, when we examine the results on AP News we can find combinations with reduced run-time (up to 70\% and 50\% reductions in train and test time respectively) with no accuracy loss (but also no gains).

\begin{figure}[b!]
    \centering
    \includegraphics[width=0.5\textwidth]{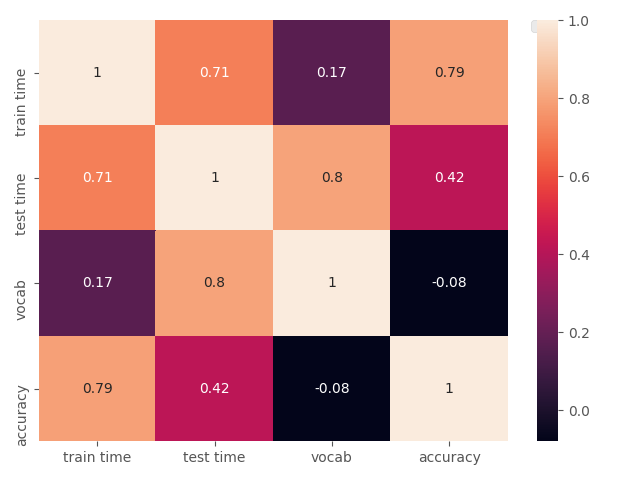}
    \caption{Pearson correlation between the relative performance variables (train time, test time, accuracy, and vocabulary size) from the results of the different preprocessing methods.}
    \label{fig:corr}
\end{figure}

\paragraph{Correlations}
In order to show the correlation between run-time and the other variables, we show a heatmap of these correlations in Figure~\ref{fig:corr}. Most of these variables are highly correlated with each other, as expected (training time is highly correlated with testing time, etc.). However, although testing time is highly correlated with vocabulary size (0.8 correlation), training time is not highly correlated (0.17), We hypothesize that a low vocabulary directly leads to faster inference, while which words are removed from the vocabulary has a bigger role in how quickly the algorithm converges during training. This hypothesis is also supported by the low correlation between vocabulary size and accuracy, indicating that what is in the vocabulary is more important than its size.

\section{Related Work}

These experiments relate to a large body of work that considers how preprocessing methods affect the downstream accuracy of various algorithms, ranging from topics in information retrieval \cite{chaudhari2015preprocessing,patil2013novel,beil2002frequent}, text classification and regression \cite{forman_extensive,Yang1997ACS,vijayarani2015preprocessing,kumar2018classification,hacohen2020influence,symeonidis_effrosynidis_arampatzis_2018,Weller2020YouDH}, topic modeling \cite{blei2003latent,Lund2019CrossreferencingUF,schofield2016comparing,schofield2017understanding,schofield_magnusson_mimno_2017}, and even more complex tasks like question answering \cite{jijkoun2003preprocessing,carvalho2007document} and machine translation \cite{habash2007syntactic,habash2006arabic,leusch2005preprocessing,weller2021streaming,mehta2020simplify} to name a few. With the rise of noisy social media, text preprocessing has become important for tasks that use data from sources like Twitter and Reddit \cite{symeonidis_effrosynidis_arampatzis_2018,singh_kumari_2016,bao_quan_wang_ren_2014,jianqiang_2015,Weller2020TheRD,zirikly2019clpsych,babanejad2020comprehensive}.

The closest lines of work to ours are those that examine how preprocessing affects text classification accuracy, where recent works like \citet{symeonidis_effrosynidis_arampatzis_2018} and \citet{hacohen2020influence} 
 analyze and cross-compare up to 16 different techniques for four machine learning algorithms.
In contrast, our work is the first to examine these preprocessing techniques beyond accuracy, examining them in tandem with how they affect vocabulary size and run-time.

\section{Conclusion}
In this work we conduct the first study that examines the relationship between vocabulary size, run-time, and accuracy across different models and corpora for text classification. 
In general, we find that although vocabulary size is highly correlated with testing time, it is not highly correlated with training time or accuracy. In these cases, the specifics of the preprocessing algorithm (the content of what it removes) matter more.

Our experiments show that rare word filtering and stopword removal are superior to many other common preprocessing methods, both in terms of their ability to reduce run-time and their potential to increase accuracy. By using these methods, we show that it is possible to reduce training and testing time by up to 65\% with a loss of only 2-5\% of accuracy, or in some cases, to provide accuracy and run-time improvements simultaneously. We hope that this study can help both researchers and industry practitioners as they design machine learning pipelines to reach their end-goals.

\bibliographystyle{acl_natbib}
\bibliography{bib}
\appendix

\section{Rare Word Filtering}
\label{app:rare}
Tables~\ref{tab:rare_amazon} and \ref{tab:rare_ap_news} show the results of rare word filtering on the Amazon and AP News datasets. We filtered at levels corresponding to the geometric progression of values from 1 to half the size of the corpus (we refer to these as levels 1 to 9, with higher numbers being more filtered).

\begin{table*}[t!]
\centering
\small
\begin{tabular}{lrrrr}
\toprule
    \# &                  Vocab Size &                   Train Time &                    Test Time &                      Accuracy \\
\midrule
  9 &  \textbf{0.0}  \textpm\ 0.0 &           \textbf{40.2}  \textpm\ 1.4 &           \textbf{51.8} \textpm\ 2.0 &           92.8 \textpm\ 0.7 \\
   8 &           0.2  \textpm\ 0.0 &           64.6  \textpm\ 2.5 &           64.0  \textpm\ 2.8 &            96.9  \textpm\ 1.1 \\
   7 &           1.0  \textpm\ 0.0 &           80.6  \textpm\ 3.0 &           70.3  \textpm\ 3.2 &   \textbf{99.3}  \textpm\ 2.8 \\
    6 &           4.4  \textpm\ 0.0 &           87.4  \textpm\ 2.7 &           72.4  \textpm\ 2.7 &  \textbf{100.3}  \textpm\ 1.8 \\
     5 &          16.4  \textpm\ 0.0 &           92.4  \textpm\ 3.6 &           76.2  \textpm\ 3.8 &   \textbf{99.7}  \textpm\ 2.4 \\
      4 &          41.4  \textpm\ 0.1 &           94.8  \textpm\ 1.9 &           83.2  \textpm\ 2.4 &  \textbf{100.0}  \textpm\ 2.1 \\
       3 &          63.3  \textpm\ 0.1 &           98.1  \textpm\ 3.2 &           90.8  \textpm\ 3.1 &   \textbf{99.7}  \textpm\ 1.8 \\
        1 &         100 \textpm\ 0.2 &          100.8  \textpm\ 2.5 &          100.1  \textpm\ 2.0 &  \textbf{100.0}  \textpm\ 1.1 \\
        2 &          77.5  \textpm\ 0.2 &          101.4  \textpm\ 5.2 &           97.1  \textpm\ 5.9 &  \textbf{100.3}  \textpm\ 1.9 \\
\bottomrule
\end{tabular}
\caption{Rare word filtering on the Amazon dataset, across various levels. Scores are the relative performance of each method over the \textit{no preprocessing} baseline. Results are the average (and std) relative performance of the four models, across the five dataset seeds. Bold indicates statistical similarity to the best score, from a two-sample t-test with $\alpha=0.05$.\label{tab:rare_amazon}}

\small
\begin{tabular}{lrrrr}
\toprule
   \# &                  Vocab Size &                   Train Time &                    Test Time &                      Accuracy \\
\midrule
 9 &  \textbf{0.0}  \textpm\ 0.0 &  \textbf{33.9}  \textpm\ 1.2 &  \textbf{61.1}  \textpm\ 3.0 &   \textbf{99.3}  \textpm\ 1.2 \\
     8 &           0.1  \textpm\ 0.0 &           51.4  \textpm\ 1.2 &  \textbf{59.1}  \textpm\ 2.2 &   \textbf{99.8}  \textpm\ 2.0 \\
  7 &           1.1  \textpm\ 0.0 &           69.6  \textpm\ 2.0 &           68.0  \textpm\ 2.0 &  \textbf{100.0}  \textpm\ 1.5 \\
   6 &           7.2  \textpm\ 0.0 &           83.8  \textpm\ 3.7 &           76.6  \textpm\ 5.7 &   \textbf{99.5}  \textpm\ 0.7 \\
    5 &          31.8  \textpm\ 0.0 &           94.0  \textpm\ 4.8 &           83.0  \textpm\ 3.7 &   \textbf{99.4}  \textpm\ 1.4 \\
     4 &          76.9  \textpm\ 0.1 &           98.2  \textpm\ 3.4 &           96.0  \textpm\ 5.5 &   \textbf{99.7}  \textpm\ 1.3 \\
       1 &         100.0  \textpm\ 0.3 &           99.4  \textpm\ 3.3 &           98.2  \textpm\ 2.7 &   \textbf{99.8}  \textpm\ 1.3 \\
       2 &          99.7  \textpm\ 0.2 &          100.3  \textpm\ 4.3 &           99.2  \textpm\ 3.9 &   \textbf{99.6}  \textpm\ 1.1 \\
      3 &          96.2  \textpm\ 0.2 &          100.6  \textpm\ 4.3 &           98.7  \textpm\ 2.4 &  \textbf{100.0}  \textpm\ 0.9 \\
\bottomrule
\end{tabular}

\caption{Rare word filtering on the AP News dataset, across various levels. This table is formatted the same as Table~\ref{tab:rare_amazon} (see the caption there for more information).\label{tab:rare_ap_news}}

\end{table*}

We find that rare word filtering at higher levels provides increased vocabulary and run-time reductions, while also reducing accuracy, in general.

\section{Word Hashing}
\label{app:hash}
Tables~\ref{tab:hash_amazon} and \ref{tab:hash_ap_news} show the effect of different levels of word hashing on model accuracy (where ``Size" indicates the number of hash buckets used). We find that word hashing with small numbers of buckets reduces vocabulary and run-time, while also decreasing accuracy in general.

\begin{table*}[hbt!]
\centering
\small
\begin{tabular}{lrrrr}
\toprule
 Size &                  Vocab Size &                   Train Time &                    Test Time &                     Accuracy \\
\midrule
   500 &  \textbf{0.1}  \textpm\ 0.0 &  \textbf{92.7}  \textpm\ 3.7 &  \textbf{76.9}  \textpm\ 3.1 &           91.7  \textpm\ 1.7 \\
 60000 &          15.1  \textpm\ 0.0 &  \textbf{96.1}  \textpm\ 1.5 &           76.7  \textpm\ 2.4 &  \textbf{99.3}  \textpm\ 1.0 \\
 10000 &           2.5  \textpm\ 0.0 &  \textbf{96.2}  \textpm\ 1.6 &  \textbf{73.2}  \textpm\ 2.2 &  \textbf{97.9}  \textpm\ 1.3 \\
  1000 &           0.3  \textpm\ 0.0 &  \textbf{96.2}  \textpm\ 4.1 &  \textbf{76.5}  \textpm\ 3.5 &           93.5  \textpm\ 1.6 \\
  40000 &          10.1  \textpm\ 0.0 &  \textbf{97.1}  \textpm\ 4.0 &  \textbf{75.7}  \textpm\ 4.0 &  \textbf{99.2}  \textpm\ 1.1 \\
  4000 &           1.0  \textpm\ 0.0 &           97.3  \textpm\ 1.1 &  \textbf{75.0}  \textpm\ 1.0 &           96.7  \textpm\ 1.6 \\
 20000 &           5.0  \textpm\ 0.0 &  \textbf{97.6}  \textpm\ 3.5 &  \textbf{74.6}  \textpm\ 3.1 &  \textbf{98.7}  \textpm\ 1.2 \\
  8000 &           2.0  \textpm\ 0.0 &  \textbf{98.2}  \textpm\ 6.8 &  \textbf{74.5}  \textpm\ 4.8 &  \textbf{97.9}  \textpm\ 1.1 \\
  2000 &           0.5  \textpm\ 0.0 &           99.2  \textpm\ 3.7 &           77.2  \textpm\ 2.6 &           95.1  \textpm\ 1.6 \\
  6000 &           1.5  \textpm\ 0.0 &          100.6  \textpm\ 6.0 &  \textbf{76.7}  \textpm\ 4.9 &           97.1  \textpm\ 1.5 \\
\bottomrule
\end{tabular}
\caption{Word Hashing on the Amazon dataset, across various levels. Scores are the relative performance of each method over the \textit{no preprocessing} baseline. Results are the average (and std) relative performance of the four models, across the five dataset seeds. Bold indicates statistical similarity to the best score, from a two-sample t-test with $\alpha=0.05$.\label{tab:hash_amazon}}

\small
\begin{tabular}{lrrrr}
\toprule
 Size &                  Vocab Size &                   Train Time &                    Test Time &                     Accuracy \\
\midrule
   500 &  \textbf{0.4}  \textpm\ 0.0 &  \textbf{93.5}  \textpm\ 2.5 &           81.4  \textpm\ 1.8 &  \textbf{99.5}  \textpm\ 1.3 \\
  1000 &           0.9  \textpm\ 0.0 &  \textbf{94.7}  \textpm\ 2.7 &  \textbf{79.9}  \textpm\ 2.6 &  \textbf{99.8}  \textpm\ 1.0 \\
  4000 &           3.5  \textpm\ 0.0 &  \textbf{94.8}  \textpm\ 2.8 &  \textbf{77.5}  \textpm\ 2.0 &  \textbf{99.4}  \textpm\ 1.2 \\
 10000 &           8.7  \textpm\ 0.0 &  \textbf{95.4}  \textpm\ 3.1 &  \textbf{78.9}  \textpm\ 3.9 &  \textbf{98.9}  \textpm\ 1.8 \\
  6000 &           5.2  \textpm\ 0.0 &  \textbf{95.6}  \textpm\ 2.9 &  \textbf{78.5}  \textpm\ 1.8 &  \textbf{99.0}  \textpm\ 0.9 \\
  2000 &           1.7  \textpm\ 0.0 &  \textbf{96.2}  \textpm\ 4.8 &  \textbf{79.9}  \textpm\ 4.0 &  \textbf{98.9}  \textpm\ 1.2 \\
 20000 &          17.3  \textpm\ 0.0 &  \textbf{96.6}  \textpm\ 3.4 &           81.2  \textpm\ 2.4 &  \textbf{99.1}  \textpm\ 1.4 \\
  8000 &           7.0  \textpm\ 0.0 &  \textbf{96.8}  \textpm\ 5.0 &  \textbf{80.3}  \textpm\ 4.7 &  \textbf{99.0}  \textpm\ 1.8 \\
  40000 &          32.8  \textpm\ 0.0 &  \textbf{98.5}  \textpm\ 4.7 &           84.7  \textpm\ 4.0 &  \textbf{99.5}  \textpm\ 1.7 \\
 60000 &          44.5  \textpm\ 0.1 &  \textbf{99.0}  \textpm\ 5.3 &           88.5  \textpm\ 6.3 &  \textbf{99.1}  \textpm\ 1.3 \\
\bottomrule
\end{tabular}
\caption{Word Hashing on the AP News dataset, across various levels. This table is formatted the same as Table~\ref{tab:hash_amazon} (see the caption there for more information).\label{tab:hash_ap_news}}
\end{table*}

\section{Full Tables for Method Combinations}
\label{app:full}
In Tables \ref{tab:full_amazon} and \ref{tab:full_ap} we show the complete table for preprocessing method combinations on Amazon and AP News respectively.

In Tables \ref{tab:knn}, \ref{tab:nb}, \ref{tab:vw}, and \ref{tab:svm}, we show the complete results for individuals models (K-NN, Naive Bayes, Vowpal Wabbit, and SVM respectively). All results are similar to the main conclusions found in the body of the paper. However, Naive Bayes in particular shows strong accuracy gains and run-time reductions for preprocessing methods, in comparison to other models.

\begin{table*}
\centering
\begin{tabular}{lrrrr}
\toprule
                        Method &                  Vocab Size &                   Train Time &                    Test Time &                      Accuracy \\
\midrule
      spell+seg+nrem+stop+rare &  \textbf{0.8}  \textpm\ 0.0 &  \textbf{44.6}  \textpm\ 1.0 &           56.8  \textpm\ 1.1 &            95.4  \textpm\ 2.0 \\
                     stop+rare &           0.9  \textpm\ 0.0 &  \textbf{46.5}  \textpm\ 3.5 &           44.5  \textpm\ 2.0 &            99.8  \textpm\ 0.8 \\
      spell+seg+nrem+stop+hash &           7.6  \textpm\ 0.0 &           53.9  \textpm\ 2.0 &  \textbf{39.6}  \textpm\ 1.4 &            97.7  \textpm\ 2.6 \\
           spell+seg+nrem+stop &          14.1  \textpm\ 0.0 &           54.2  \textpm\ 1.6 &           50.9  \textpm\ 2.3 &            97.6  \textpm\ 2.2 \\
     spell+seg+nrem+stop+lemma &          11.9  \textpm\ 0.0 &           55.1  \textpm\ 0.9 &           50.1  \textpm\ 1.7 &            97.6  \textpm\ 1.3 \\
           seg+nrem+stop+lemma &          18.5  \textpm\ 0.1 &           55.1  \textpm\ 2.3 &           54.3  \textpm\ 4.3 &            96.3  \textpm\ 3.7 \\
 spell+nopunct+nrem+stop+lemma &          31.9  \textpm\ 0.2 &           55.3  \textpm\ 1.5 &           56.7  \textpm\ 1.8 &            96.9  \textpm\ 1.2 \\
       spell+nopunct+nrem+stop &          34.1  \textpm\ 0.2 &           55.8  \textpm\ 1.6 &           56.4  \textpm\ 2.1 &            97.8  \textpm\ 1.6 \\
                 seg+nrem+stop &          21.0  \textpm\ 0.1 &           55.9  \textpm\ 1.3 &           52.5  \textpm\ 1.6 &            97.5  \textpm\ 1.4 \\
 lower+nopunct+nrem+stop+lemma &          47.0  \textpm\ 0.3 &           56.0  \textpm\ 2.2 &           61.2  \textpm\ 1.8 &            96.0  \textpm\ 1.7 \\
       lower+nopunct+nrem+stop &          48.4  \textpm\ 0.3 &           56.5  \textpm\ 1.3 &           59.4  \textpm\ 1.8 &            96.0  \textpm\ 2.3 \\
            seg+nrem+stop+stem &          14.0  \textpm\ 0.1 &           58.5  \textpm\ 1.5 &           56.8  \textpm\ 1.8 &            97.9  \textpm\ 0.7 \\
      spell+seg+nrem+stop+stem &           8.5  \textpm\ 0.0 &           58.7  \textpm\ 1.6 &           54.8  \textpm\ 2.3 &            97.8  \textpm\ 1.4 \\
  lower+nopunct+nrem+stop+stem &          39.3  \textpm\ 0.2 &           59.0  \textpm\ 1.6 &           63.2  \textpm\ 2.9 &            96.2  \textpm\ 2.4 \\
  spell+nopunct+nrem+stop+stem &          27.7  \textpm\ 0.1 &           59.1  \textpm\ 2.6 &           61.4  \textpm\ 4.0 &            97.5  \textpm\ 1.3 \\
                          stop &          99.8  \textpm\ 0.2 &           69.5  \textpm\ 1.4 &           79.0  \textpm\ 3.5 &            97.4  \textpm\ 2.2 \\
                    lower+rare &           1.0  \textpm\ 0.0 &           80.5  \textpm\ 2.0 &           69.6  \textpm\ 1.7 &   \textbf{99.8}  \textpm\ 3.1 \\
                          rare &           1.0  \textpm\ 0.0 &           80.6  \textpm\ 3.0 &           70.3  \textpm\ 3.2 &   \textbf{99.3}  \textpm\ 2.8 \\
                    spell+rare &           0.9  \textpm\ 0.0 &           80.7  \textpm\ 2.2 &           70.9  \textpm\ 2.6 &   \textbf{99.7}  \textpm\ 1.8 \\
                     stem+rare &           0.9  \textpm\ 0.0 &           81.4  \textpm\ 2.2 &           69.9  \textpm\ 1.7 &            99.4  \textpm\ 1.6 \\
                     nrem+rare &           1.0  \textpm\ 0.0 &           82.0  \textpm\ 5.6 &           70.1  \textpm\ 3.2 &            99.6  \textpm\ 1.2 \\
                  nopunct+rare &           0.9  \textpm\ 0.0 &           82.6  \textpm\ 1.9 &           70.2  \textpm\ 1.7 &  \textbf{101.0}  \textpm\ 1.8 \\
                    lemma+rare &           1.0  \textpm\ 0.0 &           82.7  \textpm\ 6.5 &           69.6  \textpm\ 1.8 &  \textbf{100.1}  \textpm\ 1.5 \\
            lower+nopunct+rare &           0.9  \textpm\ 0.0 &           86.1  \textpm\ 2.7 &           86.7  \textpm\ 2.2 &  \textbf{101.3}  \textpm\ 0.6 \\
                      seg+rare &           0.9  \textpm\ 0.0 &           86.3  \textpm\ 5.8 &           73.3  \textpm\ 3.9 &  \textbf{101.4}  \textpm\ 1.5 \\
       lower+nopunct+nrem+rare &           0.9  \textpm\ 0.0 &           87.1  \textpm\ 5.5 &           88.8  \textpm\ 5.1 &           101.1  \textpm\ 0.3 \\
                spell+seg+rare &           0.9  \textpm\ 0.0 &           89.6  \textpm\ 5.6 &           88.4  \textpm\ 5.3 &  \textbf{101.8}  \textpm\ 0.5 \\
                           seg &          24.6  \textpm\ 0.2 &           93.7  \textpm\ 2.4 &           80.3  \textpm\ 2.5 &  \textbf{100.6}  \textpm\ 1.6 \\
                         spell &          57.8  \textpm\ 0.2 &           95.1  \textpm\ 2.6 &           89.7  \textpm\ 2.6 &   \textbf{99.4}  \textpm\ 2.3 \\
                          hash &          10.1  \textpm\ 0.0 &           97.1  \textpm\ 4.0 &           75.7  \textpm\ 4.0 &            99.2  \textpm\ 1.1 \\
                       nopunct &          61.9  \textpm\ 0.2 &           97.5  \textpm\ 2.2 &           89.5  \textpm\ 2.0 &  \textbf{100.7}  \textpm\ 1.6 \\
                          stem &          81.7  \textpm\ 0.4 &           97.8  \textpm\ 2.0 &           95.0  \textpm\ 2.6 &            99.8  \textpm\ 1.0 \\
                         lower &          88.7  \textpm\ 0.3 &          101.7  \textpm\ 7.5 &          100.1  \textpm\ 6.6 &   \textbf{99.1}  \textpm\ 3.0 \\
                          nrem &          96.2  \textpm\ 0.7 &          101.7  \textpm\ 4.0 &          100.7  \textpm\ 5.3 &            99.7  \textpm\ 1.2 \\
                         lemma &          98.1  \textpm\ 0.5 &          102.2  \textpm\ 5.3 &          101.5  \textpm\ 5.1 &           100.3  \textpm\ 1.1 \\
\bottomrule
\end{tabular}
\caption{Full results of preprocessing methods on Amazon. Scores are the relative performance of each method over the \textit{no preprocessing} baseline. Results are the average (and std) relative performance of the four models, across the five dataset seeds. Bold indicates statistical similarity to the best score, from a two-sample t-test with $\alpha=0.05$.\label{tab:full_amazon}}
\end{table*}

\begin{table*}
\centering
\begin{tabular}{lrrrr}
\toprule
                        Method &                  Vocab Size &                   Train Time &                    Test Time &                      Accuracy \\
\midrule
      spell+seg+nrem+stop+rare &  \textbf{0.1}  \textpm\ 0.0 &  \textbf{29.2}  \textpm\ 0.5 &  \textbf{49.0}  \textpm\ 0.8 &   \textbf{99.3}  \textpm\ 1.7 \\
                          rare &  \textbf{0.1}  \textpm\ 0.0 &           51.4  \textpm\ 1.2 &           59.1  \textpm\ 2.2 &   \textbf{99.8}  \textpm\ 2.0 \\
       spell+nopunct+nrem+stop &          39.9  \textpm\ 0.1 &           71.0  \textpm\ 1.8 &           69.4  \textpm\ 1.8 &  \textbf{100.1}  \textpm\ 1.0 \\
 spell+nopunct+nrem+stop+lemma &          36.2  \textpm\ 0.1 &           71.3  \textpm\ 1.3 &           68.4  \textpm\ 1.4 &   \textbf{99.1}  \textpm\ 1.6 \\
           spell+seg+nrem+stop &          29.3  \textpm\ 0.0 &           72.1  \textpm\ 1.8 &           65.4  \textpm\ 3.2 &  \textbf{100.1}  \textpm\ 1.4 \\
  spell+nopunct+nrem+stop+stem &          29.6  \textpm\ 0.1 &           72.4  \textpm\ 1.2 &           68.2  \textpm\ 1.8 &   \textbf{98.7}  \textpm\ 1.8 \\
      spell+seg+nrem+stop+hash &          19.8  \textpm\ 0.0 &           73.1  \textpm\ 2.7 &           66.6  \textpm\ 3.4 &  \textbf{100.2}  \textpm\ 1.5 \\
     spell+seg+nrem+stop+lemma &          25.4  \textpm\ 0.1 &           73.6  \textpm\ 1.9 &           67.6  \textpm\ 1.7 &   \textbf{99.4}  \textpm\ 1.5 \\
 lower+nopunct+nrem+stop+lemma &          49.1  \textpm\ 0.1 &           73.7  \textpm\ 1.3 &           72.8  \textpm\ 2.0 &            98.4  \textpm\ 1.4 \\
                 seg+nrem+stop &          40.7  \textpm\ 0.2 &           74.1  \textpm\ 3.4 &           73.1  \textpm\ 3.3 &   \textbf{99.5}  \textpm\ 1.3 \\
      spell+seg+nrem+stop+stem &          19.3  \textpm\ 0.1 &           74.2  \textpm\ 2.3 &           66.9  \textpm\ 1.5 &   \textbf{99.7}  \textpm\ 1.6 \\
       lower+nopunct+nrem+stop &          51.4  \textpm\ 0.2 &           74.2  \textpm\ 3.5 &           74.0  \textpm\ 4.4 &   \textbf{99.6}  \textpm\ 1.3 \\
           seg+nrem+stop+lemma &          36.4  \textpm\ 0.1 &           75.2  \textpm\ 2.1 &           68.1  \textpm\ 3.5 &   \textbf{99.1}  \textpm\ 1.2 \\
  lower+nopunct+nrem+stop+stem &          39.4  \textpm\ 0.4 &           75.5  \textpm\ 2.4 &           72.7  \textpm\ 3.0 &  \textbf{100.3}  \textpm\ 1.0 \\
            seg+nrem+stop+stem &          29.3  \textpm\ 0.1 &           76.9  \textpm\ 2.1 &           63.9  \textpm\ 1.5 &   \textbf{99.4}  \textpm\ 1.7 \\
                          stop &          99.5  \textpm\ 0.3 &           82.5  \textpm\ 2.9 &           86.4  \textpm\ 2.0 &   \textbf{99.0}  \textpm\ 1.4 \\
                          hash &          32.8  \textpm\ 0.0 &           98.5  \textpm\ 4.7 &           84.7  \textpm\ 4.0 &   \textbf{99.5}  \textpm\ 1.7 \\
                         spell &          65.6  \textpm\ 0.2 &           98.6  \textpm\ 4.7 &           95.2  \textpm\ 7.4 &   \textbf{99.7}  \textpm\ 0.9 \\
                         lower &          92.3  \textpm\ 0.3 &           98.9  \textpm\ 1.9 &           97.6  \textpm\ 3.5 &   \textbf{99.8}  \textpm\ 1.6 \\
                          stem &          82.7  \textpm\ 0.4 &           99.1  \textpm\ 5.2 &           95.2  \textpm\ 4.0 &  \textbf{100.1}  \textpm\ 1.4 \\
                           seg &          47.5  \textpm\ 0.2 &           99.5  \textpm\ 2.4 &           88.5  \textpm\ 2.9 &  \textbf{100.3}  \textpm\ 1.3 \\
                          nrem &          89.8  \textpm\ 0.4 &           99.8  \textpm\ 3.9 &           98.5  \textpm\ 5.0 &   \textbf{99.2}  \textpm\ 1.1 \\
                       nopunct &          65.6  \textpm\ 0.2 &           99.9  \textpm\ 4.6 &           92.9  \textpm\ 4.9 &   \textbf{99.6}  \textpm\ 1.4 \\
                         lemma &          97.4  \textpm\ 0.3 &          100.5  \textpm\ 1.2 &           98.6  \textpm\ 1.6 &   \textbf{99.6}  \textpm\ 1.7 \\
\bottomrule
\end{tabular}
\caption{Full results of preprocessing methods on AP News. Scores are the relative performance of each method over the \textit{no preprocessing} baseline. Results shown are the average (and std) relative performance of the four models, across the five dataset seeds. Bold indicates statistical similarity to the best score, from a two-sample t-test with $\alpha=0.05$.\label{tab:full_ap}}
\end{table*}

\begin{table*}
\centering
\begin{tabular}{lrrrr}
\toprule
                        Method &                  Vocab Size &                   Train Time &                    Test Time &                      Accuracy \\
\midrule
      spell+seg+nrem+stop+rare &  \textbf{0.8}  \textpm\ 0.0 &  \textbf{47.0}  \textpm\ 0.8 &  \textbf{46.7}  \textpm\ 1.0 &            92.3  \textpm\ 3.3 \\
           seg+nrem+stop+lemma &          18.5  \textpm\ 0.1 &           56.7  \textpm\ 0.9 &           52.2  \textpm\ 1.6 &  \textbf{89.3}  \textpm\ 12.1 \\
 lower+nopunct+nrem+stop+lemma &          47.0  \textpm\ 0.3 &           56.8  \textpm\ 1.2 &           50.9  \textpm\ 1.0 &            90.9  \textpm\ 4.2 \\
           spell+seg+nrem+stop &          14.1  \textpm\ 0.0 &           57.0  \textpm\ 0.9 &  \textbf{48.2}  \textpm\ 1.5 &            92.2  \textpm\ 6.8 \\
     spell+seg+nrem+stop+lemma &          11.9  \textpm\ 0.0 &           57.2  \textpm\ 1.1 &           53.8  \textpm\ 0.9 &            92.8  \textpm\ 2.7 \\
      spell+seg+nrem+stop+hash &           7.6  \textpm\ 0.0 &           57.2  \textpm\ 1.4 &           48.3  \textpm\ 0.8 &            92.1  \textpm\ 6.3 \\
 spell+nopunct+nrem+stop+lemma &          31.9  \textpm\ 0.2 &           57.5  \textpm\ 1.4 &           52.9  \textpm\ 1.1 &            92.0  \textpm\ 2.2 \\
                 seg+nrem+stop &          21.0  \textpm\ 0.1 &           57.8  \textpm\ 0.7 &  \textbf{47.7}  \textpm\ 1.2 &            94.7  \textpm\ 2.5 \\
       spell+nopunct+nrem+stop &          34.1  \textpm\ 0.2 &           58.2  \textpm\ 1.6 &           48.4  \textpm\ 0.8 &            95.5  \textpm\ 3.3 \\
       lower+nopunct+nrem+stop &          48.4  \textpm\ 0.3 &           58.8  \textpm\ 1.8 &  \textbf{47.6}  \textpm\ 1.3 &            89.2  \textpm\ 6.0 \\
  lower+nopunct+nrem+stop+stem &          39.3  \textpm\ 0.2 &           59.6  \textpm\ 1.2 &           61.7  \textpm\ 4.2 &            88.0  \textpm\ 5.8 \\
  spell+nopunct+nrem+stop+stem &          27.7  \textpm\ 0.1 &           60.5  \textpm\ 2.2 &           62.8  \textpm\ 4.3 &            95.1  \textpm\ 2.5 \\
      spell+seg+nrem+stop+stem &           8.5  \textpm\ 0.0 &           61.4  \textpm\ 2.0 &           62.4  \textpm\ 4.7 &            94.2  \textpm\ 2.8 \\
            seg+nrem+stop+stem &          14.0  \textpm\ 0.1 &           61.8  \textpm\ 2.0 &           65.2  \textpm\ 0.8 &            94.2  \textpm\ 0.8 \\
                          stop &          99.8  \textpm\ 0.2 &           69.3  \textpm\ 1.8 &           52.8  \textpm\ 3.6 &   \textbf{93.4}  \textpm\ 6.6 \\
                          rare &           1.0  \textpm\ 0.0 &           80.5  \textpm\ 1.5 &           96.4  \textpm\ 0.5 &   \textbf{95.5}  \textpm\ 9.3 \\
                    lower+rare &           1.0  \textpm\ 0.0 &           82.7  \textpm\ 2.2 &          100.6  \textpm\ 2.4 &   \textbf{93.6}  \textpm\ 8.8 \\
                    spell+rare &           0.9  \textpm\ 0.0 &           82.9  \textpm\ 1.4 &          104.2  \textpm\ 6.1 &            94.0  \textpm\ 4.2 \\
                    lemma+rare &           1.0  \textpm\ 0.0 &           83.2  \textpm\ 3.2 &           99.3  \textpm\ 1.8 &   \textbf{97.4}  \textpm\ 3.5 \\
                     stem+rare &           0.9  \textpm\ 0.0 &           84.4  \textpm\ 1.9 &          101.1  \textpm\ 2.2 &            95.8  \textpm\ 3.1 \\
                  nopunct+rare &           0.9  \textpm\ 0.0 &           85.3  \textpm\ 2.7 &          101.0  \textpm\ 2.6 &   \textbf{99.4}  \textpm\ 4.4 \\
                     nrem+rare &           1.0  \textpm\ 0.0 &          91.1  \textpm\ 16.3 &          102.7  \textpm\ 8.7 &            95.6  \textpm\ 2.7 \\
                         spell &          57.8  \textpm\ 0.2 &           95.9  \textpm\ 2.7 &          103.8  \textpm\ 3.3 &   \textbf{96.4}  \textpm\ 5.7 \\
                      seg+rare &           0.9  \textpm\ 0.0 &          96.7  \textpm\ 14.5 &          112.5  \textpm\ 9.7 &  \textbf{100.6}  \textpm\ 3.5 \\
                           seg &          24.6  \textpm\ 0.2 &           98.0  \textpm\ 3.5 &          105.0  \textpm\ 2.7 &   \textbf{99.3}  \textpm\ 3.0 \\
                          stem &          81.7  \textpm\ 0.4 &           99.0  \textpm\ 1.6 &          102.4  \textpm\ 1.9 &   \textbf{98.5}  \textpm\ 2.0 \\
                         lower &          88.7  \textpm\ 0.3 &           99.4  \textpm\ 2.9 &          102.9  \textpm\ 2.8 &   \textbf{96.0}  \textpm\ 6.5 \\
                          hash &          10.1  \textpm\ 0.0 &           99.7  \textpm\ 1.3 &          100.7  \textpm\ 1.9 &   \textbf{99.4}  \textpm\ 1.2 \\
                         lemma &          98.1  \textpm\ 0.5 &          101.2  \textpm\ 2.5 &          101.9  \textpm\ 2.9 &  \textbf{100.7}  \textpm\ 1.3 \\
                       nopunct &          61.9  \textpm\ 0.2 &          101.6  \textpm\ 3.5 &          103.5  \textpm\ 3.0 &  \textbf{100.1}  \textpm\ 3.9 \\
                          nrem &          96.2  \textpm\ 0.7 &          102.1  \textpm\ 1.5 &          101.7  \textpm\ 2.4 &   \textbf{99.7}  \textpm\ 1.8 \\
\bottomrule
\end{tabular}
\caption{Effect of preprocessing techniques on Amazon with the K-NN model.  Scores are the relative performance of each method over the \textit{no preprocessing} baseline. Results shown are the average (and std) relative performance across the five dataset seeds. Bold indicates statistical similarity to the best score, from a two-sample t-test with $\alpha=0.05$.\label{tab:knn}}
\end{table*}

\begin{table*}
\centering
\begin{tabular}{lrrrr}
\toprule
                        Method &                  Vocab Size &                   Train Time &                   Test Time &                      Accuracy \\
\midrule
                     stop+rare &  \textbf{0.9}  \textpm\ 0.0 &  \textbf{44.7}  \textpm\ 5.4 &  \textbf{5.8}  \textpm\ 0.2 &  \textbf{104.5}  \textpm\ 0.8 \\
      spell+seg+nrem+stop+hash &           7.6  \textpm\ 0.0 &  \textbf{49.7}  \textpm\ 1.4 &          12.3  \textpm\ 0.0 &  \textbf{103.5}  \textpm\ 0.8 \\
           spell+seg+nrem+stop &          14.1  \textpm\ 0.0 &  \textbf{50.2}  \textpm\ 1.0 &          18.7  \textpm\ 0.6 &  \textbf{104.4}  \textpm\ 0.2 \\
     spell+seg+nrem+stop+lemma &          11.9  \textpm\ 0.0 &  \textbf{50.4}  \textpm\ 0.4 &          16.6  \textpm\ 0.3 &  \textbf{104.0}  \textpm\ 0.4 \\
           seg+nrem+stop+lemma &          18.5  \textpm\ 0.1 &           52.0  \textpm\ 2.3 &          22.7  \textpm\ 0.6 &  \textbf{103.7}  \textpm\ 0.5 \\
 spell+nopunct+nrem+stop+lemma &          31.9  \textpm\ 0.2 &           52.1  \textpm\ 0.7 &          34.8  \textpm\ 0.3 &           103.1  \textpm\ 0.9 \\
       spell+nopunct+nrem+stop &          34.1  \textpm\ 0.2 &           52.8  \textpm\ 0.8 &          36.9  \textpm\ 0.4 &           102.3  \textpm\ 1.0 \\
 lower+nopunct+nrem+stop+lemma &          47.0  \textpm\ 0.3 &           53.1  \textpm\ 1.4 &          49.1  \textpm\ 0.8 &           101.7  \textpm\ 1.4 \\
                 seg+nrem+stop &          21.0  \textpm\ 0.1 &           53.1  \textpm\ 1.6 &          25.3  \textpm\ 1.0 &  \textbf{103.6}  \textpm\ 1.7 \\
      spell+seg+nrem+stop+stem &           8.5  \textpm\ 0.0 &           53.3  \textpm\ 2.3 &          13.3  \textpm\ 0.3 &           102.7  \textpm\ 0.9 \\
  spell+nopunct+nrem+stop+stem &          27.7  \textpm\ 0.1 &           54.2  \textpm\ 0.5 &          31.1  \textpm\ 0.4 &           101.1  \textpm\ 1.4 \\
            seg+nrem+stop+stem &          14.0  \textpm\ 0.1 &           54.3  \textpm\ 0.8 &          18.7  \textpm\ 0.3 &           102.9  \textpm\ 0.7 \\
       lower+nopunct+nrem+stop &          48.4  \textpm\ 0.3 &           54.5  \textpm\ 0.7 &          50.2  \textpm\ 0.6 &           101.5  \textpm\ 1.5 \\
  lower+nopunct+nrem+stop+stem &          39.3  \textpm\ 0.2 &           55.6  \textpm\ 1.2 &          41.9  \textpm\ 0.4 &           101.5  \textpm\ 1.3 \\
                          rare &           1.0  \textpm\ 0.0 &           69.0  \textpm\ 0.8 &           6.5  \textpm\ 0.1 &           100.6  \textpm\ 1.2 \\
                    lower+rare &           1.0  \textpm\ 0.0 &           69.9  \textpm\ 1.5 &           6.4  \textpm\ 0.0 &  \textbf{102.3}  \textpm\ 1.9 \\
                          stop &          99.8  \textpm\ 0.2 &           69.9  \textpm\ 1.2 &          99.2  \textpm\ 1.2 &           101.1  \textpm\ 0.7 \\
                     nrem+rare &           1.0  \textpm\ 0.0 &           70.3  \textpm\ 1.2 &           6.5  \textpm\ 0.1 &           101.6  \textpm\ 0.9 \\
                    spell+rare &  \textbf{0.9}  \textpm\ 0.0 &           70.5  \textpm\ 0.9 &           6.4  \textpm\ 0.1 &           101.5  \textpm\ 0.9 \\
                     stem+rare &  \textbf{0.9}  \textpm\ 0.0 &           70.7  \textpm\ 1.7 &           6.4  \textpm\ 0.2 &           100.4  \textpm\ 1.7 \\
                    lemma+rare &           1.0  \textpm\ 0.0 &           71.2  \textpm\ 0.8 &           6.5  \textpm\ 0.2 &           100.7  \textpm\ 0.7 \\
                  nopunct+rare &  \textbf{0.9}  \textpm\ 0.0 &           73.2  \textpm\ 1.6 &           6.5  \textpm\ 0.2 &           101.2  \textpm\ 0.7 \\
                      seg+rare &  \textbf{0.9}  \textpm\ 0.0 &           73.3  \textpm\ 2.2 &           6.5  \textpm\ 0.1 &           101.5  \textpm\ 1.2 \\
                           seg &          24.6  \textpm\ 0.2 &           87.2  \textpm\ 1.6 &          29.1  \textpm\ 0.3 &           101.0  \textpm\ 0.7 \\
                          hash &          10.1  \textpm\ 0.0 &           90.1  \textpm\ 1.1 &          15.4  \textpm\ 0.2 &            99.3  \textpm\ 1.5 \\
                         spell &          57.8  \textpm\ 0.2 &           90.4  \textpm\ 1.7 &          59.9  \textpm\ 0.5 &            99.6  \textpm\ 1.3 \\
                       nopunct &          61.9  \textpm\ 0.2 &           94.8  \textpm\ 2.4 &          64.1  \textpm\ 0.8 &           100.6  \textpm\ 0.8 \\
                          stem &          81.7  \textpm\ 0.4 &           96.0  \textpm\ 1.0 &          82.7  \textpm\ 1.0 &            99.8  \textpm\ 1.2 \\
                         lower &          88.7  \textpm\ 0.3 &           97.7  \textpm\ 2.3 &          89.9  \textpm\ 1.5 &           100.4  \textpm\ 0.9 \\
                         lemma &          98.1  \textpm\ 0.5 &           99.5  \textpm\ 1.1 &          98.4  \textpm\ 0.9 &           100.8  \textpm\ 1.2 \\
                          nrem &          96.2  \textpm\ 0.7 &           99.8  \textpm\ 1.5 &          97.0  \textpm\ 1.6 &            99.6  \textpm\ 1.2 \\
\bottomrule
\end{tabular}
\caption{Effect of preprocessing techniques on Amazon with the Naive Bayes model. Scores are the relative performance of each method over the \textit{no preprocessing} baseline. Results shown are the average (and std) relative performance across the five dataset seeds. Bold indicates statistical similarity to the best score, from a two-sample t-test with $\alpha=0.05$.\label{tab:nb}}
\end{table*}

\begin{table*}
\centering
\begin{tabular}{lrrrr}
\toprule
                        Method &                  Vocab Size &                   Train Time &                    Test Time &                      Accuracy \\
\midrule
      spell+seg+nrem+stop+rare &  \textbf{0.8}  \textpm\ 0.0 &  \textbf{42.1}  \textpm\ 1.5 &  \textbf{46.6}  \textpm\ 1.4 &            97.0  \textpm\ 1.3 \\
                     stop+rare &           0.9  \textpm\ 0.0 &           47.7  \textpm\ 4.0 &  \textbf{51.4}  \textpm\ 4.1 &            97.5  \textpm\ 0.9 \\
       lower+nopunct+nrem+stop &          48.4  \textpm\ 0.3 &           53.3  \textpm\ 1.6 &           56.1  \textpm\ 1.2 &            97.5  \textpm\ 0.8 \\
           spell+seg+nrem+stop &          14.1  \textpm\ 0.0 &           54.3  \textpm\ 3.2 &           57.2  \textpm\ 3.3 &            97.9  \textpm\ 0.9 \\
 spell+nopunct+nrem+stop+lemma &          31.9  \textpm\ 0.2 &           54.4  \textpm\ 2.3 &           57.9  \textpm\ 2.7 &            97.3  \textpm\ 1.2 \\
           seg+nrem+stop+lemma &          18.5  \textpm\ 0.1 &           54.5  \textpm\ 4.0 &           59.9  \textpm\ 9.5 &            97.3  \textpm\ 1.2 \\
 lower+nopunct+nrem+stop+lemma &          47.0  \textpm\ 0.3 &           54.9  \textpm\ 5.7 &           58.0  \textpm\ 4.8 &            96.7  \textpm\ 0.7 \\
       spell+nopunct+nrem+stop &          34.1  \textpm\ 0.2 &           54.9  \textpm\ 3.1 &           57.8  \textpm\ 2.8 &            97.9  \textpm\ 0.8 \\
      spell+seg+nrem+stop+hash &           7.6  \textpm\ 0.0 &           55.0  \textpm\ 3.0 &           58.3  \textpm\ 3.4 &            97.5  \textpm\ 0.7 \\
                 seg+nrem+stop &          21.0  \textpm\ 0.1 &           55.4  \textpm\ 2.3 &           58.5  \textpm\ 2.5 &            97.2  \textpm\ 0.8 \\
            seg+nrem+stop+stem &          14.0  \textpm\ 0.1 &           55.5  \textpm\ 1.8 &           57.8  \textpm\ 2.0 &            98.1  \textpm\ 0.6 \\
      spell+seg+nrem+stop+stem &           8.5  \textpm\ 0.0 &           57.2  \textpm\ 0.8 &           59.6  \textpm\ 0.9 &            98.1  \textpm\ 0.9 \\
  lower+nopunct+nrem+stop+stem &          39.3  \textpm\ 0.2 &           57.2  \textpm\ 3.1 &           59.9  \textpm\ 3.2 &            98.4  \textpm\ 1.4 \\
  spell+nopunct+nrem+stop+stem &          27.7  \textpm\ 0.1 &           58.9  \textpm\ 5.9 &           61.9  \textpm\ 4.6 &            98.1  \textpm\ 0.5 \\
                          stop &          99.8  \textpm\ 0.2 &           66.9  \textpm\ 1.5 &           69.2  \textpm\ 2.4 &            97.9  \textpm\ 0.7 \\
                     nrem+rare &           1.0  \textpm\ 0.0 &           78.5  \textpm\ 3.5 &           79.1  \textpm\ 3.4 &  \textbf{100.6}  \textpm\ 0.6 \\
                    spell+rare &           0.9  \textpm\ 0.0 &           79.1  \textpm\ 3.6 &           79.7  \textpm\ 2.8 &  \textbf{101.7}  \textpm\ 1.1 \\
                    lemma+rare &           1.0  \textpm\ 0.0 &           79.8  \textpm\ 4.4 &           80.0  \textpm\ 3.9 &  \textbf{101.5}  \textpm\ 1.0 \\
                     stem+rare &           0.9  \textpm\ 0.0 &           79.9  \textpm\ 3.3 &           80.2  \textpm\ 3.0 &  \textbf{100.8}  \textpm\ 0.8 \\
                    lower+rare &           1.0  \textpm\ 0.0 &           80.2  \textpm\ 2.8 &           80.5  \textpm\ 2.3 &  \textbf{101.7}  \textpm\ 0.7 \\
                  nopunct+rare &           0.9  \textpm\ 0.0 &           80.5  \textpm\ 2.2 &           80.7  \textpm\ 2.1 &  \textbf{101.8}  \textpm\ 1.2 \\
            lower+nopunct+rare &           0.9  \textpm\ 0.0 &           80.9  \textpm\ 3.6 &           81.7  \textpm\ 3.5 &  \textbf{101.4}  \textpm\ 0.7 \\
                      seg+rare &           0.9  \textpm\ 0.0 &           82.9  \textpm\ 4.8 &           83.4  \textpm\ 5.1 &  \textbf{101.8}  \textpm\ 0.5 \\
                          rare &           1.0  \textpm\ 0.0 &           84.1  \textpm\ 9.0 &           84.9  \textpm\ 8.3 &  \textbf{100.6}  \textpm\ 0.3 \\
       lower+nopunct+nrem+rare &           0.9  \textpm\ 0.0 &           84.4  \textpm\ 9.5 &           85.7  \textpm\ 9.1 &  \textbf{101.3}  \textpm\ 0.2 \\
                spell+seg+rare &           0.9  \textpm\ 0.0 &           86.1  \textpm\ 9.4 &           86.9  \textpm\ 8.9 &  \textbf{102.1}  \textpm\ 0.5 \\
                           seg &          24.6  \textpm\ 0.2 &           94.5  \textpm\ 1.9 &           94.6  \textpm\ 2.6 &  \textbf{102.2}  \textpm\ 1.4 \\
                       nopunct &          61.9  \textpm\ 0.2 &           95.3  \textpm\ 2.6 &           95.8  \textpm\ 2.0 &  \textbf{101.9}  \textpm\ 0.5 \\
                         spell &          57.8  \textpm\ 0.2 &           98.6  \textpm\ 5.0 &           98.1  \textpm\ 6.0 &  \textbf{101.5}  \textpm\ 1.0 \\
                          stem &          81.7  \textpm\ 0.4 &           98.9  \textpm\ 4.6 &           97.7  \textpm\ 4.3 &  \textbf{100.9}  \textpm\ 0.5 \\
                          hash &          10.1  \textpm\ 0.0 &         102.8  \textpm\ 12.2 &         102.1  \textpm\ 12.3 &  \textbf{100.5}  \textpm\ 0.8 \\
                          nrem &          96.2  \textpm\ 0.7 &         104.3  \textpm\ 11.9 &         103.3  \textpm\ 11.4 &           100.0  \textpm\ 0.8 \\
                         lower &          88.7  \textpm\ 0.3 &         107.9  \textpm\ 17.2 &         107.7  \textpm\ 15.5 &  \textbf{100.9}  \textpm\ 1.7 \\
                         lemma &          98.1  \textpm\ 0.5 &         109.4  \textpm\ 16.3 &         107.8  \textpm\ 14.9 &           100.0  \textpm\ 1.0 \\
\bottomrule
\end{tabular}
\caption{Effect of preprocessing techniques on Amazon with the Vowpal Wabbit model.  Results shown are the average (and std) relative performance, across the five dataset seeds. Bold indicates statistical similarity to the best score, from a two-sample t-test with $\alpha=0.05$.\label{tab:vw}}
\end{table*}

\begin{table*}
\centering
\begin{tabular}{lrrrr}
\toprule
                        Method &                  Vocab Size &                   Train Time &                    Test Time &                      Accuracy \\
\midrule
      spell+seg+nrem+stop+rare &  \textbf{0.8}  \textpm\ 0.0 &  \textbf{44.7}  \textpm\ 0.6 &  \textbf{77.1}  \textpm\ 0.9 &            96.7  \textpm\ 1.4 \\
                     stop+rare &           0.9  \textpm\ 0.0 &           47.1  \textpm\ 1.1 &  \textbf{76.2}  \textpm\ 1.6 &            97.4  \textpm\ 0.8 \\
           spell+seg+nrem+stop &          14.1  \textpm\ 0.0 &           55.4  \textpm\ 1.5 &  \textbf{79.3}  \textpm\ 3.8 &            95.7  \textpm\ 0.7 \\
           seg+nrem+stop+lemma &          18.5  \textpm\ 0.1 &           57.1  \textpm\ 1.8 &  \textbf{82.2}  \textpm\ 5.6 &            94.9  \textpm\ 1.0 \\
       spell+nopunct+nrem+stop &          34.1  \textpm\ 0.2 &           57.2  \textpm\ 0.8 &           82.7  \textpm\ 4.4 &            95.7  \textpm\ 1.1 \\
                 seg+nrem+stop &          21.0  \textpm\ 0.1 &           57.4  \textpm\ 0.5 &  \textbf{78.5}  \textpm\ 1.7 &            94.5  \textpm\ 0.4 \\
 spell+nopunct+nrem+stop+lemma &          31.9  \textpm\ 0.2 &           57.4  \textpm\ 1.5 &           81.4  \textpm\ 3.2 &            95.1  \textpm\ 0.6 \\
     spell+seg+nrem+stop+lemma &          11.9  \textpm\ 0.0 &           57.6  \textpm\ 1.3 &  \textbf{80.1}  \textpm\ 3.9 &            96.1  \textpm\ 0.9 \\
 lower+nopunct+nrem+stop+lemma &          47.0  \textpm\ 0.3 &           59.3  \textpm\ 0.6 &           86.7  \textpm\ 0.6 &            94.8  \textpm\ 0.5 \\
       lower+nopunct+nrem+stop &          48.4  \textpm\ 0.3 &           59.5  \textpm\ 1.3 &           83.7  \textpm\ 4.1 &            95.7  \textpm\ 0.9 \\
            seg+nrem+stop+stem &          14.0  \textpm\ 0.1 &           62.6  \textpm\ 1.3 &           85.7  \textpm\ 4.2 &            96.6  \textpm\ 0.6 \\
      spell+seg+nrem+stop+stem &           8.5  \textpm\ 0.0 &           62.7  \textpm\ 1.2 &           84.1  \textpm\ 3.2 &            96.2  \textpm\ 1.1 \\
  spell+nopunct+nrem+stop+stem &          27.7  \textpm\ 0.1 &           62.8  \textpm\ 1.6 &           89.8  \textpm\ 6.8 &            95.9  \textpm\ 0.7 \\
  lower+nopunct+nrem+stop+stem &          39.3  \textpm\ 0.2 &           63.5  \textpm\ 0.9 &           89.5  \textpm\ 4.0 &            96.8  \textpm\ 1.3 \\
                          stop &          99.8  \textpm\ 0.2 &           71.9  \textpm\ 1.3 &           94.6  \textpm\ 6.9 &            97.3  \textpm\ 0.8 \\
                     nrem+rare &           1.0  \textpm\ 0.0 &           88.3  \textpm\ 1.5 &           92.1  \textpm\ 0.7 &           100.4  \textpm\ 0.5 \\
                          rare &           1.0  \textpm\ 0.0 &           88.7  \textpm\ 0.9 &           93.6  \textpm\ 4.0 &           100.5  \textpm\ 0.5 \\
                    lower+rare &           1.0  \textpm\ 0.0 &           89.1  \textpm\ 1.4 &           90.7  \textpm\ 1.8 &  \textbf{101.5}  \textpm\ 0.8 \\
       lower+nopunct+nrem+rare &           0.9  \textpm\ 0.0 &           89.9  \textpm\ 1.5 &           92.0  \textpm\ 1.2 &  \textbf{100.9}  \textpm\ 0.3 \\
                    spell+rare &           0.9  \textpm\ 0.0 &           90.3  \textpm\ 2.8 &           93.3  \textpm\ 1.5 &  \textbf{101.5}  \textpm\ 1.2 \\
                     stem+rare &           0.9  \textpm\ 0.0 &           90.5  \textpm\ 2.0 &           91.9  \textpm\ 1.6 &  \textbf{100.7}  \textpm\ 0.7 \\
                  nopunct+rare &           0.9  \textpm\ 0.0 &           91.3  \textpm\ 1.1 &           92.6  \textpm\ 1.9 &  \textbf{101.6}  \textpm\ 0.8 \\
            lower+nopunct+rare &           0.9  \textpm\ 0.0 &           91.3  \textpm\ 1.9 &           91.8  \textpm\ 0.8 &  \textbf{101.2}  \textpm\ 0.6 \\
                      seg+rare &           0.9  \textpm\ 0.0 &           92.4  \textpm\ 1.9 &           90.9  \textpm\ 0.7 &  \textbf{101.6}  \textpm\ 0.6 \\
                spell+seg+rare &           0.9  \textpm\ 0.0 &           93.2  \textpm\ 1.8 &           89.9  \textpm\ 1.6 &  \textbf{101.5}  \textpm\ 0.6 \\
                           seg &          24.6  \textpm\ 0.2 &           95.0  \textpm\ 2.6 &           92.7  \textpm\ 4.5 &  \textbf{100.0}  \textpm\ 1.5 \\
                         spell &          57.8  \textpm\ 0.2 &           95.6  \textpm\ 1.2 &           97.0  \textpm\ 0.8 &            99.9  \textpm\ 1.2 \\
                          hash &          10.1  \textpm\ 0.0 &           95.6  \textpm\ 1.4 &           84.7  \textpm\ 1.5 &            97.7  \textpm\ 0.8 \\
                    lemma+rare &           1.0  \textpm\ 0.0 &          96.9  \textpm\ 17.6 &           92.5  \textpm\ 1.4 &  \textbf{101.0}  \textpm\ 0.7 \\
                          stem &          81.7  \textpm\ 0.4 &           97.3  \textpm\ 0.8 &           97.4  \textpm\ 3.2 &            99.9  \textpm\ 0.2 \\
                       nopunct &          61.9  \textpm\ 0.2 &           98.5  \textpm\ 0.4 &           94.8  \textpm\ 2.1 &           100.2  \textpm\ 1.0 \\
                         lemma &          98.1  \textpm\ 0.5 &           98.7  \textpm\ 1.5 &           97.9  \textpm\ 1.8 &            99.8  \textpm\ 0.8 \\
                          nrem &          96.2  \textpm\ 0.7 &          100.8  \textpm\ 1.2 &          100.9  \textpm\ 5.9 &            99.5  \textpm\ 1.0 \\
\bottomrule
\end{tabular}
\caption{Effect of preprocessing techniques on Amazon with the SVM model. Scores are the relative performance of each method over the \textit{no preprocessing} baseline.  Results shown are the average (and std) relative performance across the five dataset seeds. Bold indicates statistical similarity to the best score, from a two-sample t-test with $\alpha=0.05$.\label{tab:svm}}
\end{table*}

\end{document}